# Predictive Analytics in E-Commerce for Customer Behavior Forecasting using hybrid Ret-DNN with XGBoost Model

1st Degala Pushpa Sri
*Department of MBA*
*Pace Institute of Technology and Sciences*
Valluru, Ongole, Andhra Pradesh, India
pushpasri.degala@gmail.com

2nd Mayank Atreya
*Chewy Inc*
Atlanta, USA
mayankatreya2@gmail.com

3rd Lakshmi. H
*Department of Management Studies*
*Nitte Meenakshi Institute of Technology, Nitte (Deemed to be University)*
Bengaluru, India
lakshmi.h@nmit.ac.in

4th Navin Chhibber
*Infinity Tech Group*
Sunnyvale, CA, USA
naveenchibber.research@gmail.com

5th Mukesh Soni
*Division of Research and Development*
*Lovely Professional University*
Phagwara, India
mukesh.research24@gmail.com

***Abstract*—In recent years, electronic (E) - commerce services have rapidly increased in the daily lives of people, which helps them to purchase products online. However, retail platforms have struggled to understand customer behavior and make it difficult to predict their future purchases. To overcome these challenges, this study proposes a hybrid Retail Deep Neural Network (Ret-DNN) with an Extreme Gradient Boosting (XGBoost) model for capturing temporal features and tabular dynamics of retail data. First, data were sourced from a United Kingdom (UK)-based online retailer that contains transactions with almost 500,000 records. Then, the collected data were pre-processed using a series of techniques, such as data cleaning, outlier handling, temporal feature extraction, feature encoding, and z-score normalization, to ensure that the data were ready for model training and testing. Subsequently, the preprocessed data were fed into the Ret-DNN model, which acts as a feature extractor to understand the complete context of customer transactions. Further, the extracted data were fed as input into the XGBoost model, which predicted the final output as the purchase probability of customers. Finally, the proposed Ret-DNN XGBoost model achieved better results by attaining a Mean Absolute Error (MAE) 0.2193 when compared to the existing Ret-DNN model.**

## I. INTRODUCTION

Over the past few years, usage of E-commerce services has been rapidly increased globally in daily lives of people for so many essential reasons such as retailing, shopping, online banking, bill payments, online marketing and so many. E-commerce is an online platform that provides goods to sell and purchase, including digital money transactions through the Internet and other electronic networks [1]. Moreover, e-commerce platforms have the potential to support businesses by suggesting products based on their previous purchases, which eventually improves their business market [2]. Additionally, these platforms can easily identify the churn rate, which provides an idea about customer engagement and the credibility of their platform for customers. Furthermore, their platforms analyze customer opinions from their reviews and social media interactions to fine-tune their offers according to market demand [3]. However, in the past, most e-commerce companies relied on manual reports and spreadsheets for forecasting trends, which is a time-consuming process that takes weeks and months. Additionally, these methods are based on simple averages or linear regression techniques, which cannot completely understand market trends [4]. Moreover, these methods are mostly able to provide results that have happened, but are not able to predict what will happen, as business will be able to react only to churn, but could not prevent it.

In addition, these methods cannot handle data from multiple resources such as social media, bill transactions, and IOT devices [5]. To overcome these challenges, Big Data Analytics (BDA) was introduced for advanced Machine Learning (ML) models and predictive clustering methods, which have the ability to automate data analysis based on different kinds of data, thereby saving weeks of manual work [6]. Correspondingly, Business Intelligence (BI) tools, such as tableau or power BI, were utilized for cleaning and processing data and for further data visualization dashboards that clearly show an overview of business growth. Additionally, churn detection utilizes ML techniques to classify customers in real time engagement, such as churn, non-churn, active, and inactive, which helps businesses prevent customer loss [7]. However, the ML algorithm has a high risk of overfitting when handling noisy data.

The state of art methods include decision tree and random forest models were introduced for splitting customers data on certain conditions such as payment history and browsing time. Moreover, these two combinations improve stability and accuracy, making the segmentation process easy, such as churn risk vs. loyal customers [8]. Simultaneously, Clustering methods have been introduced to group similar groups of customers for advanced market analysis. Moreover, these methods support recommended systems by suggesting products based on their credibility in the market [9]. Furthermore, these clustering-based models can understand temporal features, which makes it difficult to handle real-time data. However, these predictive analytics access large amounts of data from various customers, which may lead to privacy and security concerns [10].

The main contribution of this research:

- The hybrid Ret-DNN XGBoost model is proposed to understand customer behavior forecasting for E Commerce using predictive analytics while reducing computational cost without compromising accurate performance.
- Then, the collected data were preprocessed using a series of techniques, such as z-score normalization, data cleaning, outlier handling, and handling data imbalance issues.
- Subsequently, the preprocessed data were fed into the Retail Deep Neural Network (Ret-DNN) model to extract features for further feeding them as inputs to the XGBoost model.

The remainder of this paper is organized as follows: Section 2 specifies the literature review, Section 3 demonstrates the proposed methodology, Section 4 explains the experimental results with a corresponding discussion, and Section 5 presents the conclusion.

## II. LITERATURE REVIEW

Rifat Al Mamun Rudro et al. [11] presented predictive analytics for customer behavior analysis in retail, using the Ret-DNN model. Initially, data were collected from e-commerce transaction logs, which contained features including the time of purchase, prices, product categories, and customer purchase. The first collected data were preprocessed by removing duplicates and null values, and later normalization was performed for scaling the numerals. Later, the preprocessed data were fed into the Retail Deep Neural Network (Ret-DNN) model to understand the complex relationship of customer behavior.

Moreover, this model has the potential to update business stays according to the market trends. In addition, this model easily understands the intent of future customer purchases. However, this model requires the continuous monitoring of relevant trends, which is a difficult challenge. Alghanam et al. [12] presented a data-mining model to understand the e-commerce context for predicting customer purchase behavior. Initially, the data were sourced from a publicly available database consisting of customer details and product categories.

The collected data were preprocessed using normalization for feature scaling, and all null and missing values were removed to avoid inaccurate model performance. Subsequently, the preprocessed data were fed into the k-means algorithm model to create groups that are clusters based on similarity. Subsequently, different decision tree algorithms were applied to predict customer behavior.

Moreover, this model can automate recommendations on Amazon or Flipkart shopping websites. Additionally, these models have the ability to group various types of customers, such as frequent buyers, occasional buyers, and discount seekers. However, this model struggles to adapt to fast-changing, real-life patterns. Xiancheng Xiahou and Yoshio Harada [13] demonstrated an e-commerce customer churn prediction system using K-means and SVM models. Initially, data were sourced from a dataset published by the Alibaba Cloud Tianchi platform, which contains the behavioral data of 987,994 users. Additionally, regarding their shopping events in 2017, features included behavior type, timestamp, User ID and Item ID.

The collected data were then cleaned by fixing the time stamps with the proper time slots AM and PM. Later, the preprocessed data were fed into the k-means algorithm for customer segmentation, and the SVM algorithm was applied for churn prediction. Moreover, this model provides churn prediction, which helps identify customer losses. However, this model requires many cloud servers, which incur high implementation costs.

Md. Ahmmed et al. [14] demonstrated enhanced deep learning-based customer demand prediction for understanding e-commerce dynamics. Initially, data were sourced from the Amazon review dataset in 2018, which consists of ratings for various product categories provided by different customers. First-source data were preprocessed using min max normalization for scaling into a uniform manner, and Term Frequency Inverse Term Frequency (TF-IDF) was performed for word and sentence tokenization. Later, preprocessed data were fed into several models, such as the Conditional Transformer Language Model (CTRL), Long Short-Term Memory (LSTM), and Gated Recurrent Unit (GRU) models for comparative evaluation, where CTRL outperformed other models with better metrics such as F1-score, precision, accuracy, and recall.

Moreover, the CTRL model has the ability to predict customer demand for grocery products, which reduces expired waste food. Additionally, this model can capture meaning in the text reviews of customers after purchasing their products. However, this model training requires many hardware components, such as a GPU, which leads to high computational costs. Sayyad et al. [15] demonstrated a predictive modeling framework for optimizing e-commerce supply chains with categorical boosting. Initially, data were sourced from the DataCo Global dataset, which contains features, including carrier details, delivery dates, product details, order dates, and product ID.

The first collected data were preprocessed by removing duplicate values, fixing null values, and detecting outliers. Subsequently, the preprocessed data were fed into the CatBoost algorithm for regression purposes by selecting relevant features. Moreover, this framework can identify customer complaints as early as possible, instead of reacting after customer complaints. In addition, this model has the potential to address customer complaints as early as possible. However, training this CatBoost algorithm requires high numbers of CPUs, GPUs, and expensive cloud servers, which increases the computational cost.

## III. METHODOLOGY

In this study, a hybrid Ret-DNN XGBoost model is proposed to understand customer behavior forecasting for E Commerce using predictive analytics, while reducing computational cost without compromising accurate performance. Initially, data were sourced by a United Kingdom (UK) based online retailer, which consists of customer transactions of almost 500,000 records from various European countries. Then, the collected data were preprocessed using a series of techniques, such as z-score normalization, data cleaning, outlier handling, and handling data imbalance issues. Subsequently, the preprocessed data were fed into the Ret-DNN model to extract features for further feeding them as inputs to the XGBoost model. Finally, this hybrid architecture compresses all numerical features to reduce dimensionality, which eventually balances the computational efficiency and predicts the final output. The overall methodology of the proposed Ret-DNN XGBoost model is explained in subsequent sections and is shown in the below Figure 1.

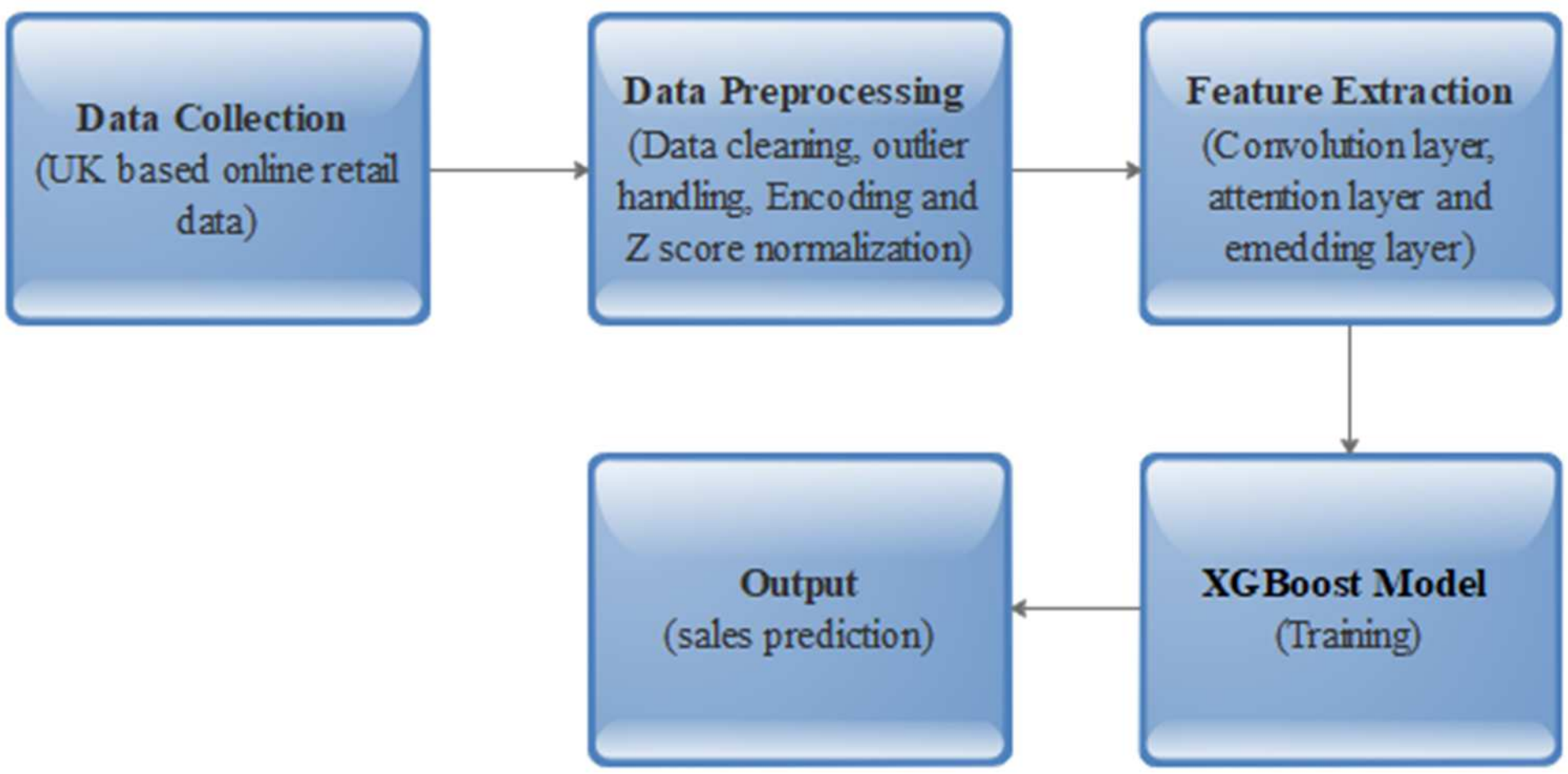

Fig. 1. Pictorial representation of proposed Ret-DNN XGBoost

## A. Data Collection

In this research, the United Kingdom (UK) [11]-based online retail non-store data was considered as a standard dataset to understand customer behavior, which contains transactions of customers from December 1, 2010, to December 9, 2011. Additionally, this data set contains approximately 500,000 records of almost 95% from the UK and rest from other European countries France, Germany, and Spain. Furthermore, these dataset features include InvoiceNo, CustomerID, StockCode, Quantity, Unitprice, and Country. Each feature describes detailed information of customers such as InvoiceNo refers to nominal six-digit transactions, CustomerID refers to five-digit numbers, StockCode refers to five-digit product identifiers, Quantity refers to the number of quantities for each product per transaction, Unitprice refers to each price unit of a continuous product, and Country refers to the country name to which each customer belongs. Furthermore, the collected data were preprocessed using various techniques, as demonstrated in the subsequent section.

## B. Data preprocessing

In this section, the collected data were preprocessed using a series of techniques. First, the data were cleaned by applying logarithmic transformation to handle outliers, and duplicate invalid records were removed from each customer transaction, such as money refunded or cancelled transactions, which are not useful for predicting sales and avoiding any misreading of model prediction. Correspondingly, resensy-based temporal feature extraction was applied to convert InvoiceDate to a date time format, such as hour, day, month, and last customer purchase time. Subsequently, frequency encoding and inverse frequency weighting techniques were performed for high-cardinality categories and correcting class imbalance issues. Finally, the Z-score normalization technique was applied for feature scaling to all numerical features, such as unit price and quantity. All the techniques and equations used for data preprocessing with detailed explanations are shown below in (1) – (5).

*1) Data cleaning:* In data cleaning, all invalid records in the dataset were removed by USING k-Nearest-Neighbor imputation (KNN) to avoid biases and unnecessary deviations in model prediction for predicting sales. Moreover, this ensures that further preprocessing steps are performed smoothly using only valid data. The equation used for this data preprocessing step is shown in Equation (1).

$$D_{clean} = \{x_i \in D \mid Quantity_i > 0,\ UnitPrice_i > 0, Invoice\ No_i \notin C\} \quad (1)$$

Where $D$ denotes the original dataset, $D_{\text{clean}}$ denotes the cleaned dataset, Quantity $_i$ denotes the total number of items purchased in transaction $i$, Quantity $_i$ denotes the transaction $i$ how many items purchased, Unit Price $_i$ denotes transaction $i$ 's price per unit of a product, and $C$ denotes the number of refunded invoice numbers.

*2) Outlier handling:* Outlier handling is performed after data cleaning, which controls the effect of extreme purchase quantities and stabilizes the variance for unusually high transactions using fuzzy matching. This technique was performed using Equation (2).

$$x' = log(1 + x) \quad (2)$$

Where $x$ represents the original value and $x'$ represents the transformed value. Additionally, adding 1 yields the results for $x = 0$, which ensures that all numerical features have a certain distributed range.

*3) Temporal Feature Extraction:* After handling outliers, temporal feature extraction was applied to convert the InvoiceDate attribute into meaningful time format values using 1D-CNN. This technique allows the model to identify cyclic-based trends because all customer purchase patterns are time-dependent. The equation for this technique is given in Equation (3).

$$Recency_c = Date_{ref} - LastPurchaseDate_c \quad (3)$$

Where $Date_{ref}$ denotes the reference date, $LastPurchaseDate_c$ denotes Customer $C$ 's most recent purchase date, and $Recency_c$ denotes the total number of days since the last purchase.

*4) Encoding categorical variables and Imbalance Correction:* The encoding technique was applied after temporal feature extraction to convert categorical features into numerical features. correspondingly, imbalance correction assures occasional item contributions in model learning. Equations (4)–(5) are used for both the encoding and class imbalance.

Where $f(c_i)$ denotes category's frequency, $w_i$ denotes instance $i$'s sample weight, $\epsilon$ denotes constant and $\alpha$ denotes monitoring weight on rare classes which belongs to the values between [0.5,1.0].

$$Enc(c_i) = \frac{f(c_i)}{\sum_j f(c_j)} \quad (4)$$

$$w_i = \frac{1}{(f(c_i)+\epsilon)^{\alpha}} \quad (5)$$

*5) Feature scaling using Z Score Normalization:* The Z Score normalization technique is applied before feeding into the Ret-DNN model, where it scales all numerical values into uniform values. This technique improves the model convergence and avoids instability.

$$z_i = \frac{x_i - \mu}{\sigma} \quad (6)$$

As shown in Equation (6), where $x_i$ denotes a unique feature value, $\mu$ denotes the mean value of the feature, $\sigma$ denotes the standard deviation of the feature, and $z_i$ denotes the standardized feature value. This ensures that all values have a mean of zero and a standard deviation of one.

## C. Model Architecture

In this stage, the Ret-DNN model is trained with the preprocessed data, where it acts as a feature extractor that extracts temporal features and understands the context from the preprocessed retail transaction data. Moreover, this architecture does not increase the computational cost while maintaining an efficient performance. This architecture follows a series of steps, including a convolution layer, recurrent layer, attention layer, and an embedding layer. Correspondingly, the detailed architectural steps are explained in the subsequent sections.

*1) Convolution layer:* In convolution layer, after preprocessing and dataset splitting into 80 % for training, 10 % for validation, and 10% for testing. Subsequently, all the input sequences are $X_t = \{x_1, x_2, \dots, x_T\}$ where these sequences represent recent transactions of $T$ . This convolution layer captures the short-term dependencies of transaction window Equation (7).

$$h_t^{(c)} = ReLU(W_c * X_t + b_c) \quad (7)$$

Where $W_c$ denotes convolution kernel, $*=$ denotes convolution operator and $b_c$ is bias.

*2) Recurrent layer:* After the convolution layer, the local patterns are extracted through the recurrent layer to aggregate the local signals based on the transaction timestamps. This Gated Recurrent Unit (GRU) identifies future customer behavior based on previous purchases. The equation used for this recurrent layer as mentioned below Equation (8).

Where $f_{GRU}$ denotes the unit function of gated recurrent, and $h_t^{(r)}$ hidden state of recurrent.

$$h_t^{(r)} = f_{GRU}\left(h_{t-1}^{(r)}, h_t^{(c)}\right) \quad (8)$$

*3) Attention layer:* This layer helps the model to understand the context by focusing on the most relevant time steps, such as only important purchases for the prediction. The equation used for this layer is given below Equation (9).

$$\alpha_t = \frac{e^{u_t}}{\sum_{k=1}^{T} e^{u_k}}, u_t = v^T tanh\left(W_a h_t^{(r)} + b_a\right) \quad C = \sum_{t=1}^{T} \alpha_t h_t^{(r)} \quad (9)$$

Where $\alpha_t$ denotes the $t$'s attention score of step t, and $C$ denotes the final context vector for customer behavior.

*4) Embedding projection layer:* Finally, after the attention layer, this embedding layer projects the context vector $C$ into a low-dimensional embedding, which reduces the computational cost. Further, the output of this layer acts as an input to the XGBoost meta-learner. The equation for this embedding layer is mentioned below Equation (10).

$$e_i = W_e C + b_e \quad (10)$$

Where $e_i \in \mathbb{R}^d$ denotes the embedding of sample $i$ 's embedding in dimension $d$ and $W_e$ denotes the matrix weight in the embedding space. Hence, the output of the last embedding layer before classification is a high-dimensional feature vector — i.e., deep learned embeddings that capture the semantic patterns of retail data (product features, sales patterns, etc.).

## D. XGBoost Meta-Learner

In this stage, the Extreme Gradient Boosting (XGBoost) model uses all inputs extracted from the final embedding layer of the Ret-DNN architecture. First, each vector refers to a compressed feature set of transactional customer behaviors. This model learns non-linear patterns and handles imbalanced data while maintaining a low computational cost compared to other deep neural networks. Moreover, this model was trained to represent transactions without overfitting problems. The equations used in this model are given by Equations (11)–(12).

$$\hat{y}_i = \sum_{k=1}^{K} f_k(e_i), f_k \in \mathcal{F} \quad (11)$$

$$\mathcal{L}(\phi) = \sum_{i=1}^{N} l(y_i, \hat{y}_i) + \sum_{k=1}^{K} \Omega(f_k) \quad (12)$$

Where $l(y_i, \hat{y}_i)$ denotes the loss function, $\hat{y}_i$ refers to the predicted value of $i$, $e_i$ refers to the embedding vector, $f_k$ regression value, and $\mathcal{F}$ refers the space in all the regression values.

## E. Final prediction

In this final prediction phase after completion of training, the XGBoost meta – learner predicts the final output. This hybrid integration allows the model to perform with less computational cost and is extremely accurate for prediction. Furthermore, this integration of gradient boosting with extraction controls residual transaction errors, even in low-frequency transactions. In addition, this architecture ensures predictive robustness, computational efficiency, and fast and efficient deployment. The equation used to predict the final output is mentioned below Equation (13).

$$\hat{y}_{new} = XGBoost\left(RetDNN(X_{new})\right) \quad (13)$$

Where $X_{\text{new}}$ refers to customer data which was pre-processed, RetDNN is a feature extractor, and $XGBoost$ refers to the final output predictor. The XGBoost meta-learner uses the deep feature embeddings created by the Ret-DNN as input and performs the final prediction. It learns compound non-linear relationships to improve accuracy and generalization. The final output is either a predicted class label (e.g., demand level or product category) or a continuous value (e.g., sales forecast) which are clearly demonstrated in the following section.

# IV. Experimental Results

In this study, the Ret-DNN XGBoost model was proposed to predict customer behavior in retail sales by integrating deep neural networks and gradient boosting optimization techniques. This hybrid model could identify both sequential patterns and non-linear relationships while ensuring robust predictive accuracy without compromising the computational cost. To implement this Python3.10 and Deep learning libraries such as TensorFlow, XGBoost, and keras were run on a system with an Intel I5 processor in Windows 11 with 16 GB RAM. The evolution metric equations are given below Equations (14)–(16).

$$Mean\ Absolute\ Error\ (MAE) = \frac{1}{n}\sum_{i=1}^{n} |y_i - \hat{y}_i| \quad (14)$$

$$Root\ Mean\ Square\ Error\ (RMSE) = \sqrt{\frac{1}{n}\sum_{i=1}^{n} (y_i - \hat{y}_i)^2} \quad (15)$$

$$Accuracy = \frac{TP+TN}{TP+TN+FP+FN} \quad (16)$$

Where $n$ denotes the number of observations, $y_i$ denotes the target variable of $i^{th}$ observation, $\hat{y}_i$ is the predicted value, and $|y_i - \hat{y}_i|$ is the difference between the predicted and the actual values.

## A. Performance Analysis

In the performance analysis, the proposed Ret-DNN XGBoost hybrid model was evaluated using traditional models such as CNN-LSTM, GRU-Attention, and Hybrid Fuzzy Neural Network to evaluate its effectiveness for forecasting retail sales.

In this performance analysis, the proposed model outperformed the traditional models because of the efficient integration of temporal feature extraction with gradient boosting tree-based learning. As a result, the proposed Ret-DNN XGBoost model achieved superior results, as shown in Figures 2 and 3.

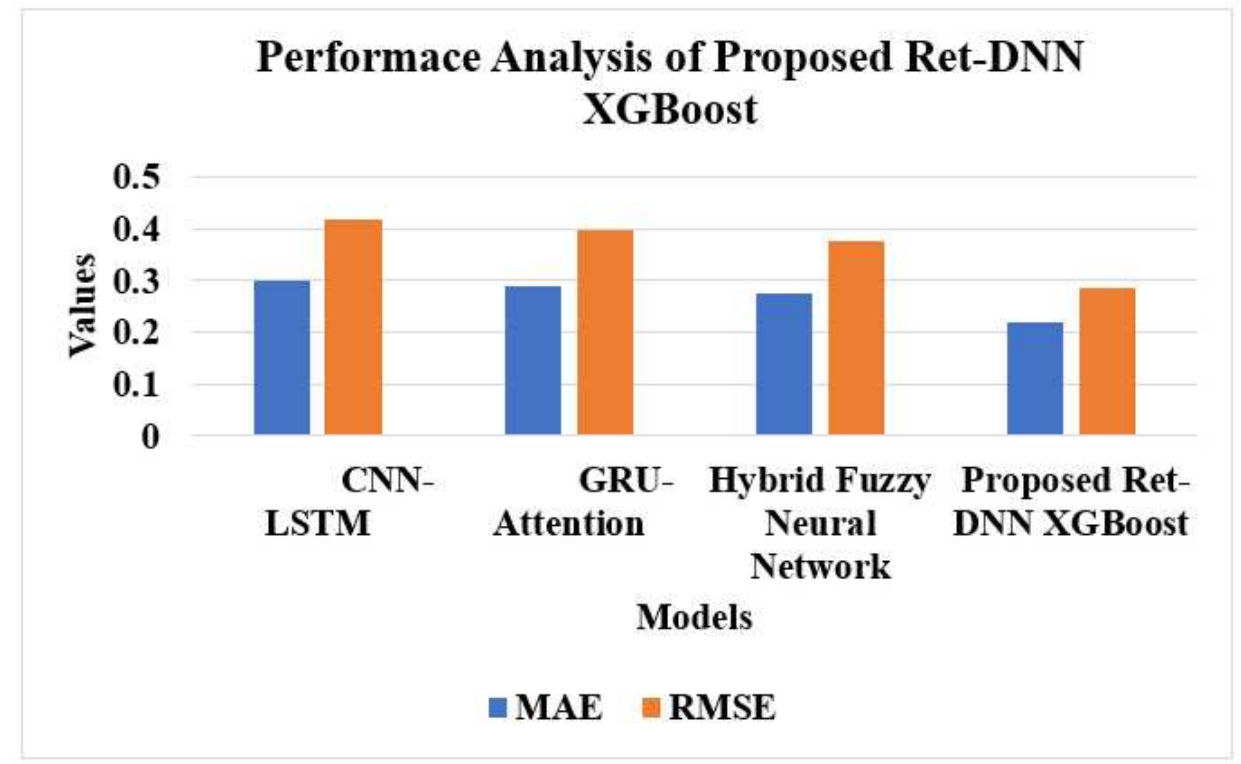

Fig. 2. Performance analysis of proposed Ret-DNN XGBoost in terms of MAE and RMSE

From Figures 2 and 3, it is clear that the proposed Ret-DNN XGBoost model obtained better results than the traditional CNN-LSTM, GRU-Attention, and Hybrid Fuzzy Neural Network models with a Mean Average Error (MAE) of 0.2193, Root Mean Square Error (RMSE) of 0.2858, and accuracy of 0.92. CNN-LSTM with MAE of 0.2976, RMSE of 0.4158, and accuracy of 0.86; GRU-attention with MAE of 0.2880, RMSE of 0.3958, and accuracy of 0.84; and Hybrid Fuzzy Neural Network with MAE of 0.2761, RMSE of 0.3758, and accuracy of 0.81, respectively.

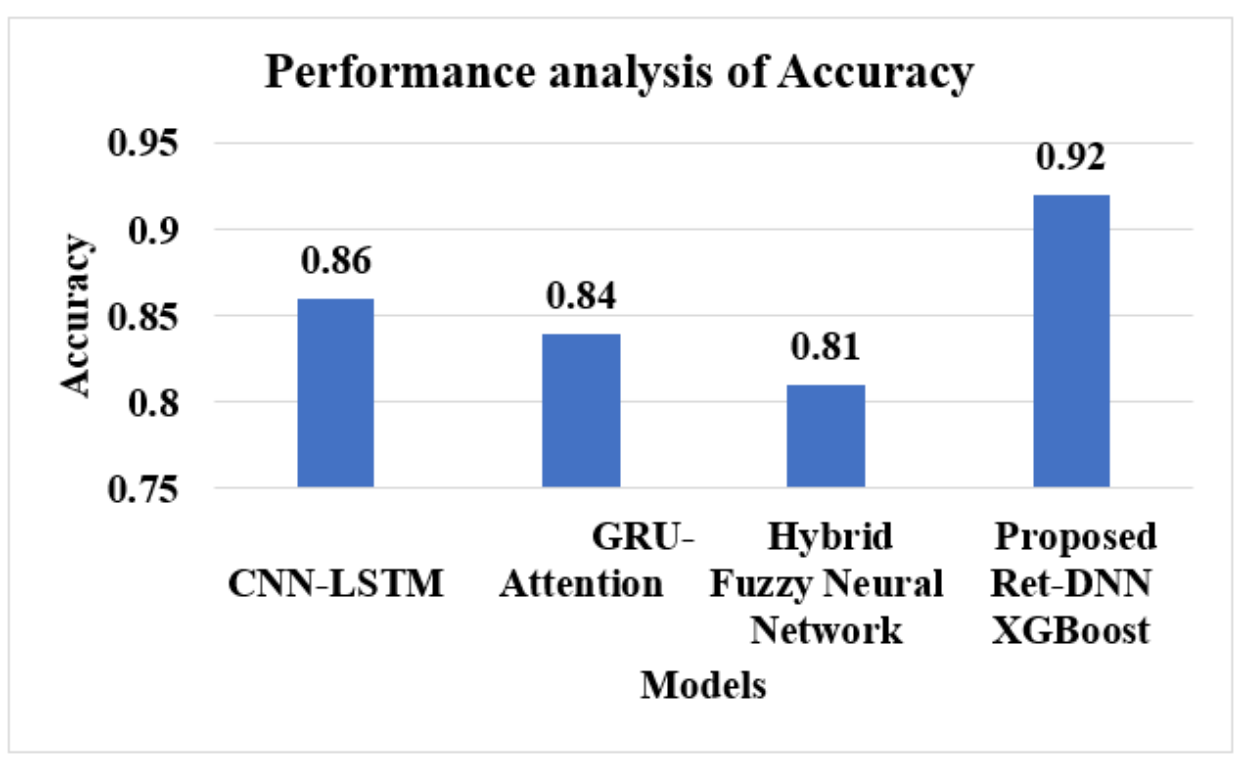

Fig. 3. Performance analysis of proposed Ret-DNN XGBoost model in terms of accuracy

## B. Comparative Analysis

The efficiency of the improved performance of the proposed Ret-DNN XGBoost model was evaluated using the existing Ret-DNN model, and compared with the considered performance metrics.

Therefore, the proposed model outperformed the existing Ret-DNN model by achieving a better MAE, RMSE, Accuracy, as shown in Table 1.

TABLE I. COMPARATIVE ANALYSIS OF PROPOSED RET-DNN XGBOOST

| Models | MAE | RMSE | Accuracy |
|---|---|---|---|
| Ret DNN [11] | 0.2539 | 0.3508 | 0.91 |
| Proposed Ret-DNN XGBoost | 0.2193 | 0.2858 | 0.95 |

From Table 1, it is clear that the proposed Ret-DNN XGBoost model obtained better results than the existing model, with an MAE of 0.2193, RMSE of 0.2858, and accuracy of 0.95. In contrast, Ret-DNN [11] had an MAE of 0.2539, RMSE of 0.3508, and accuracy of 0.91.

### C. Discussion

The primary objective of this study is to develop predictive analytics in e-commerce using Ret-DNN XGBoost to understand customer behavior for forecasting sales. The results show that the proposed Ret-DNN XGBoost overcomes the challenges of the existing Ret DNN model and traditional models, such as CNN-LSTM, GRU-Attention, and Hybrid Fuzzy Neural Network, to evaluate its effectiveness for forecasting retail sales. The existing Ret DNN [11] model was unable to maintain computational efficiency and capture the temporal features of customer transactions. Thus, the proposed Ret-DNN XGBoost temporal features can predict residual transaction errors even at low transaction frequencies. Moreover, this model has the potential to recognize insights into unseen data, such as complex and dynamic sales patterns. Similarly, CNN-LSTM, GRU-Attention and Hybrid Fuzzy Neural Network models fail to identify nonlinear dependencies, such as the seasonal behavior of a customer. In this research, the proposed Ret-DNN XGBoost Model integrates a gradient boosting technique that improves optimization and stabilizes performance in real-world retail environments. Furthermore, this model has the potential to capture both long- and short-term dependencies, such as understanding time-based customer behavior, which traditional models fail to capture. Finally, this proposed Ret-DNN XGBoost model decisively outperformed the Ret DNN CNN-LSTM, GRU-Attention, and hybrid fuzzy neural network models by building a hybrid architecture to predict customer behavior for forecasting sales in e-commerce.

## V. Conclusion

This research paper demonstrated predictive analytics for e-commerce to understand customer behavior for forecasting sales using the proposed Ret-DNN XGBoost Model. The traditional model challenges were resolved using the Ret-DNN XGBoost Model by integrating deep neural networks and a gradient boosting technique for forecasting retail sales. Initially, data were collected from UK based online retail, which contains all the transactional and purchase data of customers. Then, the collected data were preprocessed using various techniques, such as data cleaning, outlier handling, temporal feature extraction, feature encoding, handling data imbalance, and z-score normalization. Later, the preprocessed data were processed into a Ret-DNN model, which acted as a feature extraction tool. This Ret-DNN model is processed through a convolution layer, an attention layer, and an embedding layer, and all features such as temporal information and purchasing behavior are extracted. Subsequently, the output of the Ret-DNN model was used as the input to train the XGBoost model to predict the final output as the purchase probability of a customer. Furthermore, the results clearly illustrate that the proposed Ret-DNN XGBoost model obtained better results than the existing model, with an MAE of 0.2193 and RMSE of 0.2858. In the future, the proposed model will further explore Federated Learning (FL) which ensures privacy and security while handling e-commerce data from multiple sources.